\documentclass[11pt,a4paper]{article}
\usepackage[hyperref]{acl2019}
\usepackage{times}
\usepackage{latexsym}
\usepackage{amsmath}  

\usepackage{booktabs}
\usepackage{mathtools}
\usepackage{txfonts}

\usepackage{algorithm}
\usepackage{algorithmic}
\usepackage{graphicx}
\usepackage{enumitem}
\usepackage{multirow}
\usepackage{caption}
\usepackage{csquotes}
\usepackage{mathabx}
\usepackage[normalem]{ulem}
\usepackage{xcolor,colortbl, makecell}
\usepackage{multicol}
\usepackage{xcolor}
\usepackage{soul}
\usepackage{framed}

\newcommand{\hlc}[2][yellow]{{%
    \colorlet{foo}{#1}%
    \sethlcolor{foo}\hl{#2}}%
}

\newcommand{\correctans}[1]{\hlc[cyan!30]{\textbf{#1}}}
\newcommand{\incans}[1]{\hlc[pink!80]{#1}}
\newcommand{\missed}[1]{\hlc[black!10]{\textbf{#1}}}

\newcommand{\aquestion}[6]{\begin{tabular}{p{0.1cm}p{6.6cm}}
\multicolumn{2}{l}{\parbox{6.9cm}{\underline{#1} \\ #2 \\}} \\ 
& #3\\ 
& #4\\
& #5\\
& #6\\
\end{tabular}
}
\newcommand\swagtwofont[1]{{\usefont{T1}{damion}{m}{n}#1}}
\newcommand\swagtwofonttitle[1]{{{\usefont{T1}{damion}{m}{n}#1}}}
\newcommand\com[1]{}

\usepackage[colorinlistoftodos,prependcaption,textsize=tiny]{todonotes}

\newcommand{\olddatasetname}{{SWAG}}

\newcommand{\newdatasetname}{{\swagtwofont{HellaSwag}}}
\newcommand{\newdatasetnamelong}{{\swagtwofont{H}}arder {\swagtwofont{E}}ndings, {\swagtwofont{L}}onger contexts, and {\swagtwofont{L}}ow-shot {\swagtwofont{A}}ctivities for {\swagtwofont{S}}ituations {\swagtwofont{W}}ith {\swagtwofont{A}}dversarial {\swagtwofont{G}}enerations}
\newcommand{\newdatasetnametitle}{{\swagtwofonttitle{HellaSwag}}}

\usepackage[colorinlistoftodos,prependcaption,textsize=tiny]{todonotes}
\newcommand{\secsubtitle}[1]{}

\usepackage{url}

\aclfinalcopy 

\title{
\newdatasetnametitle: 
Can a Machine \emph{Really} Finish Your Sentence? 
}

\author{Rowan Zellers$^\spadesuit$ \: \: 
  Ari Holtzman$^{\spadesuit}$ \: \: 
  \bf Yonatan Bisk$^\spadesuit$ \: \:
  Ali Farhadi$^{\spadesuit\heartsuit}$ \: \:
  Yejin Choi$^{\spadesuit\heartsuit}$\\
  $^\spadesuit$Paul G. Allen School of Computer Science \& Engineering, University of Washington \\
  $^\heartsuit$Allen Institute for Artificial Intelligence\\
  \vspace{1mm} \url{https://rowanzellers.com/hellaswag}
  }

\date{}

\begin{document}
\maketitle
\pagestyle{plain}
\thispagestyle{plain}
\begin{abstract}
Recent work by \citet{zellers2018swagaf} introduced a new task of \emph{commonsense natural language inference}: given an event description such as ``A woman sits at a piano,'' a machine must select the most likely followup: ``She sets her fingers on the keys.'' With the introduction of BERT \cite{devlin2018bert}, near human-level performance was reached. Does this mean that machines can perform human level commonsense inference?

In this paper, we show that commonsense inference still proves difficult for even state-of-the-art models, by presenting \newdatasetname, a new challenge dataset. Though its questions are trivial for humans (${>}$95\% accuracy), state-of-the-art models struggle (${<}$48\%). We achieve this via Adversarial Filtering (AF), a data collection paradigm wherein a series of discriminators iteratively select an adversarial set of machine-generated wrong answers. AF proves to be surprisingly robust. The key insight is to scale up the length and complexity of the dataset examples towards a critical `Goldilocks' zone wherein generated text is ridiculous to humans, yet often misclassified by state-of-the-art models.

Our construction of \newdatasetname, and its resulting difficulty, sheds light on the inner workings of deep pretrained models. More broadly, it suggests a new path forward for NLP research, in which benchmarks co-evolve with the evolving state-of-the-art in an adversarial way, so as to present ever-harder challenges.
\end{abstract}

\section{Introduction}
\begin{figure}[t!]
  \centering\small
    \includegraphics[width=1\columnwidth]{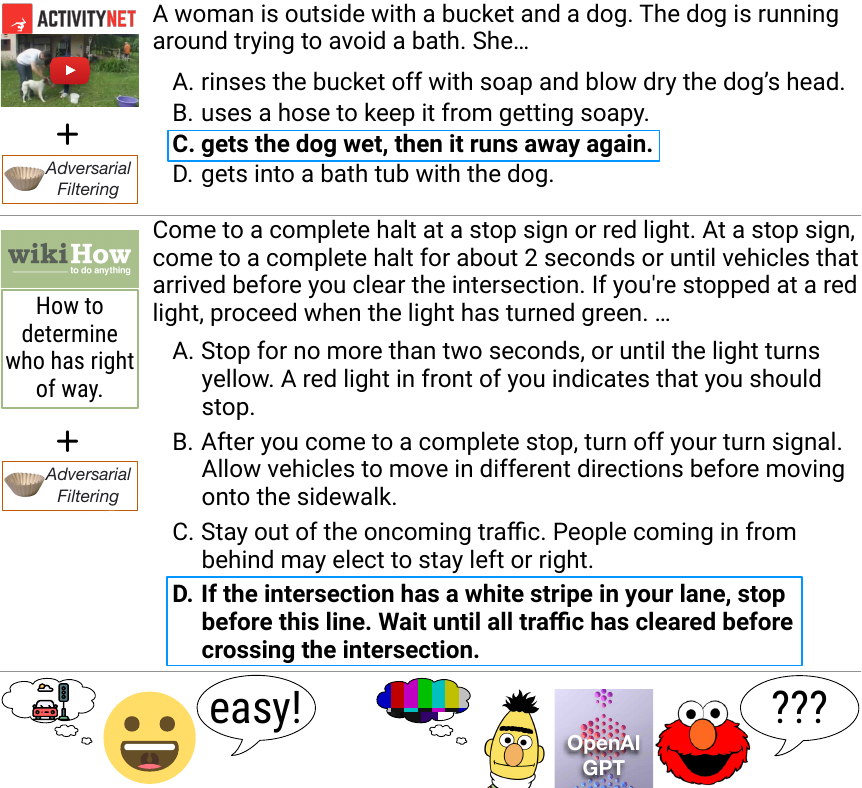}
    \vspace{-1mm}
\caption{Models like BERT struggle to finish the sentences in \newdatasetname, even when they come from the same distribution as the training set.
While the wrong endings are on-topic, with words that relate to the context, humans consistently judge their meanings to be either incorrect or implausible. For example, option \textbf{A} of the WikiHow passage suggests that a driver should stop at a red light for \textbf{no \underline{more} than two seconds}.
}
  \label{fig:teaser}
\end{figure}
Imagine a woman chasing a dog around outside, trying to give it a bath. What might happen next? Humans can read a narrative like this, shown in Figure~\ref{fig:teaser}, and connect it to a rich model of the world: the dog is currently dry and not soapy, and it actively doesn't want to be bathed. Thus, one plausible next event is option \textbf{C}---that she'll get the dog wet and it will run away again. 


When the SWAG
dataset was first announced \cite{zellers2018swagaf}, this new task of \emph{commonsense natural language inference} seemed trivial for humans (88\%) and yet challenging for then-state-of-the-art models (${<}$60\%), including ELMo \cite{peters2018deep}. However, BERT \cite{devlin2018bert} soon reached over 86\%, almost human-level performance. One news article on this development was headlined 
``\emph{finally, a machine that can finish your sentence.}''\footnote{A New York Times article at \href{https://nyti.ms/2DycutY}{https://nyti.ms/2DycutY}.}

In this paper, we investigate the following question: How well do deep pretrained models, like BERT, perform at commonsense natural language inference (NLI)? Our surprising conclusion is that the underlying \emph{task} remains unsolved. Indeed, we find that deep models such as BERT do not demonstrate robust commonsense reasonining ability by themselves. Instead, they operate more like \emph{rapid surface learners} for a particular dataset. Their strong performance on \olddatasetname{} is dependent on the finetuning process, wherein they largely learn to pick up on dataset-specific distributional biases.
When the distribution of language shifts slightly, performance drops drastically -- even if the domain remains identical.

We study this question by introducing  \newdatasetname,\footnote{Short for \newdatasetnamelong. Dataset and code at \href{https://rowanzellers.com/hellaswag}{https://rowanzellers.com/hellaswag}.} a new benchmark for commonsense NLI. We use Adversarial Filtering (AF), a data-collection paradigm in which a series of discriminators is used to select a challenging set of generated wrong answers. AF is surprisingly effective towards this goal: the resulting dataset of 70k problems is easy for humans (95.6\% accuracy), yet challenging for machines (${<}50\%)$. This result holds even when models are given a significant number of training examples, and even when the test data comes from the exact same distribution as the training data. Machine performance slips an additional 5\% when evaluated on examples that cover novel concepts from the same domain.

To make this dataset robust to deep pretrained models, we use a trifecta of state-of-the-art generators \cite{radford2018improving}, state-of-the-art discriminators (BERT), and high quality source text. We expand on the SWAG's original video-captioning domain by using WikiHow articles, greatly increasing the context diversity and generation length. Our investigation reveals a Goldilocks zone -- roughly three sentences of context, and two generated sentences -- wherein generations are largely nonsensical, even though state-of-the-art discriminators cannot reliably tell the difference between these generations and the ground truth. 

More broadly, our paper presents a case-study towards a future of verified progress in NLP, via iterative rounds of building and breaking datasets. If our ultimate goal is to provide reliable benchmarks for challenging tasks, such as commonsense NLI, these benchmarks cannot be static. Instead, they must evolve together with the evolving state-of-the-art. Continued evolution in turn requires principled dataset creation algorithms. Whenever a new iteration of a dataset is created, these algorithms must leverage existing modeling advancements to filter out spurious biases. Only once this cycle becomes impossible can we say that the underlying \emph{task} -- as opposed an individual dataset -- is solved. 
\section{Background}


\olddatasetname~is a dataset for commonsense NLI. For each question, a model is given a \textbf{context} from a video caption and four \textbf{ending choices} for what might happen next. Only one choice is right -- the actual next caption of the video.

Obtaining interesting negatives is challenging. Prior work (e.g. \citealp{gururangan2018annotation,poliak_hypothesis_2018}) has found that when humans write the endings to NLI questions, they introduce subtle yet strong class-conditional biases known as \emph{annotation artifacts}.\footnote{These biases simply inflate model performance, but past work has also shown that are unwanted social biases induced when humans write the endings, in terms of gender and race \cite{rudinger_learning_2015}.}

\begin{figure}[t!]
  \centering\small
    \includegraphics[width=\columnwidth]{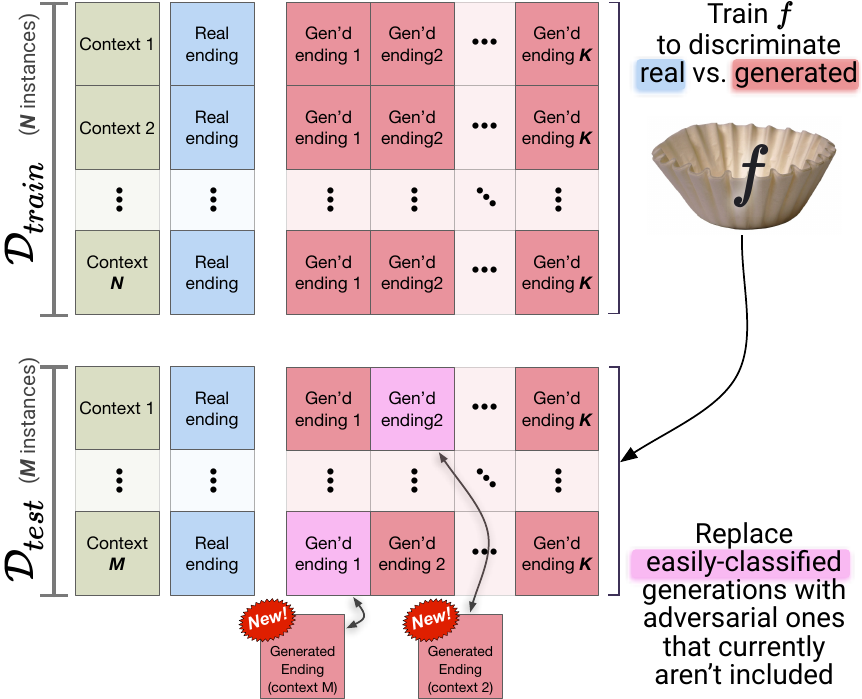}
\caption{An overview of Adversarial Filtering. On each iteration, a new classifier is trained on a dummy training set $\mathcal{D}_{train}$ to replace easily-classified negative endings on the dummy test set $\mathcal{D}_{test}$ with adversarial endings. This process is repeated iteratively, to obtain a challenging dataset regardless of the final split.}
  \label{fig:afdiagram}
\end{figure}

To address this, \citet{zellers2018swagaf} introduced \textbf{Adversarial Filtering} (AF). An overview is shown in Figure~\ref{fig:afdiagram}. The key idea is to produce a dataset $\mathcal{D}$ which is adversarial for \emph{any} arbitrary split of $(\mathcal{D}_{train}$, $\mathcal{D}_{test})$. 
This requires a \emph{generator} of negative candidates (i.e., wrong endings that violate human notions about how the world works), which we achieve by using a language model. 
Potential candidates of incorrect answers were massively oversampled from a language model trained on in-domain data, and then selected using an ensemble of adversaries. The selection process happens iteratively: on each iteration, the dataset is randomly partitioned into $\mathcal{D}_{train}$ and $\mathcal{D}_{test}$. The ensemble is trained to classify endings as \emph{real} or \emph{generated} on $\mathcal{D}_{train}$, then, AF replaces easy-to-classify generations in $\mathcal{D}_{test}$. This process continues until the accuracy of these adversaries converges. Last, humans validate the data to remove adversarial endings that seem realistic.

Importantly, AF creates a final dataset that is challenging to models regardless of the final dataset split. In Section~\ref{sec:hellaswag}, we will use AF as the underlying workhorse to construct an NLI dataset that is easy for humans, yet challenging for machines. This difficulty persists even when models are provided significant training data, and even when this data comes from the same distribution as the test set. This contrasts with past work on adversarial examples (e.g.  \citealp{jia2017adversarial, glockner2018breaking, belinkov2018noise}) which consider cases where an out-of-distribution test set is constructed to be adversarial.

\section{Investigating SWAG}
\begin{figure}[t!]
  \centering\small
    \includegraphics[width=\columnwidth]{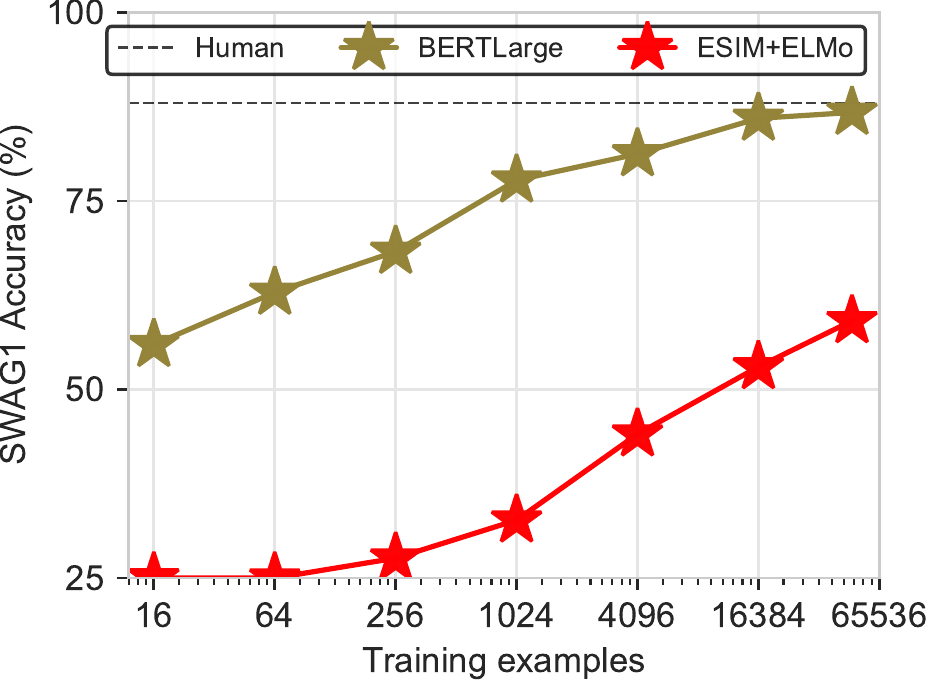}
\caption{Validation accuracy on~\olddatasetname~for BERTLarge versus training set size. The baseline (25\% accuracy) is random chance. BERT does well given as few as 16 training examples, but requires tens of thousands of examples to approach human performance.}
  \label{fig:learningcurve}
\end{figure}

In this section, we investigate why \olddatasetname~was solved. We focus on BERT, since it is the best known approach at the time of writing.\footnote{See the appendix for a discussion of the BERT architecture and hyperparameter settings we used in our experiments.} Core to our analysis is investigating how a model trained on Wikipedia and books can be so effectively finetuned for SWAG, a dataset from video captions. 

\subsection{How much innate knowledge does BERT have about SWAG?}

We investigate this question by measuring BERT's performance on \olddatasetname~while varying the size of the training dataset; results are shown in Figure~\ref{fig:learningcurve}. While the best known ELMo NLI model (ESIM+ELMo; \citealp{chen2017enhanced}) requires the entire training set to reach 59\%, BERT outperforms this given only 64 examples. However, BERT still needs upwards of 16k examples to approach human performance, around which it plateaus.

\subsection{What is learned during finetuning?}

\begin{figure}[t!]
  \centering\small
    \includegraphics[width=\columnwidth]{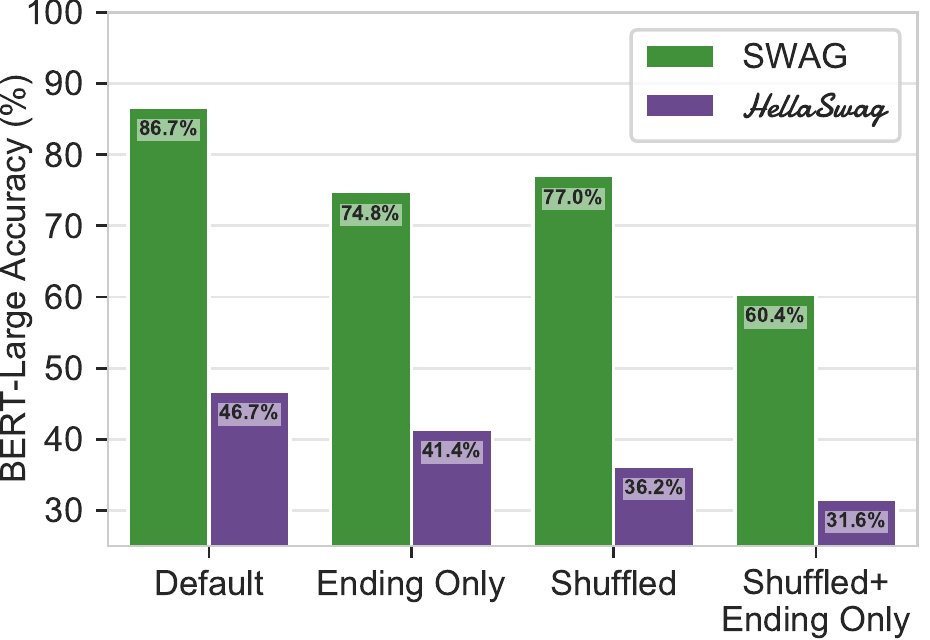}
    \caption[]{BERT validation accuracy when trained and evaluated under several versions of \olddatasetname, with the new dataset \newdatasetname~as comparison. We compare:}
    \begin{small}
    \begin{tabular}{@{\hspace{0pt}}p{6em}@{\hspace{3pt}}p{17em}@{}}
    \texttt{Ending Only}& No context is provided; just the endings.\\[2pt]%
    \texttt{Shuffled}& Endings that are indidivually tokenized, shuffled, and then detokenized.\\[2pt]
    \texttt{Shuffled+ Ending Only}& No context is provided \emph{and} each ending is shuffled.
    \end{tabular}
    \end{small}
  \label{fig:ablations}
\end{figure}

\begin{figure*}
  \centering\small
    \includegraphics[width=0.48\textwidth]{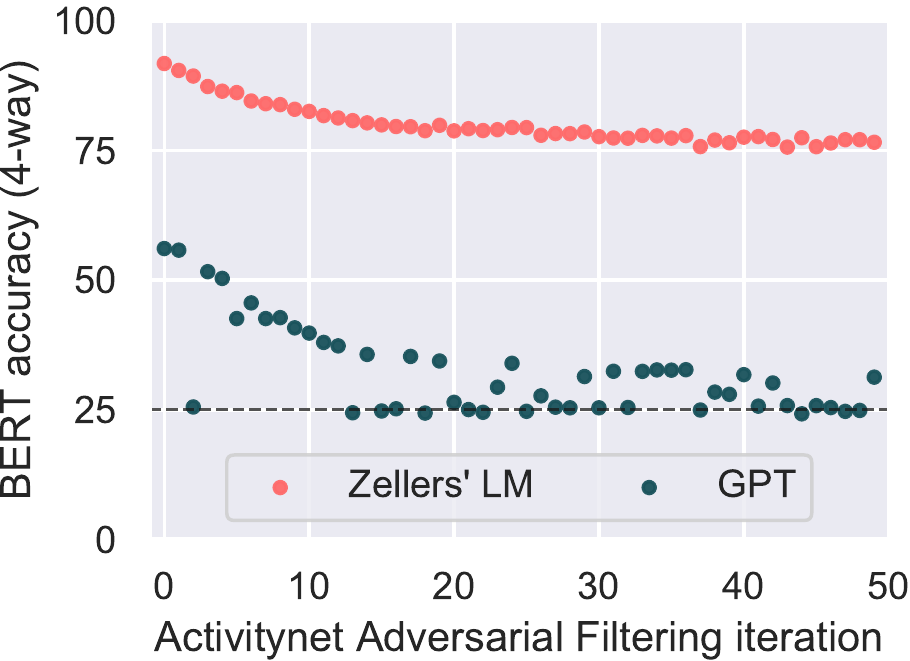}\hfill
    \includegraphics[width=0.48\textwidth]{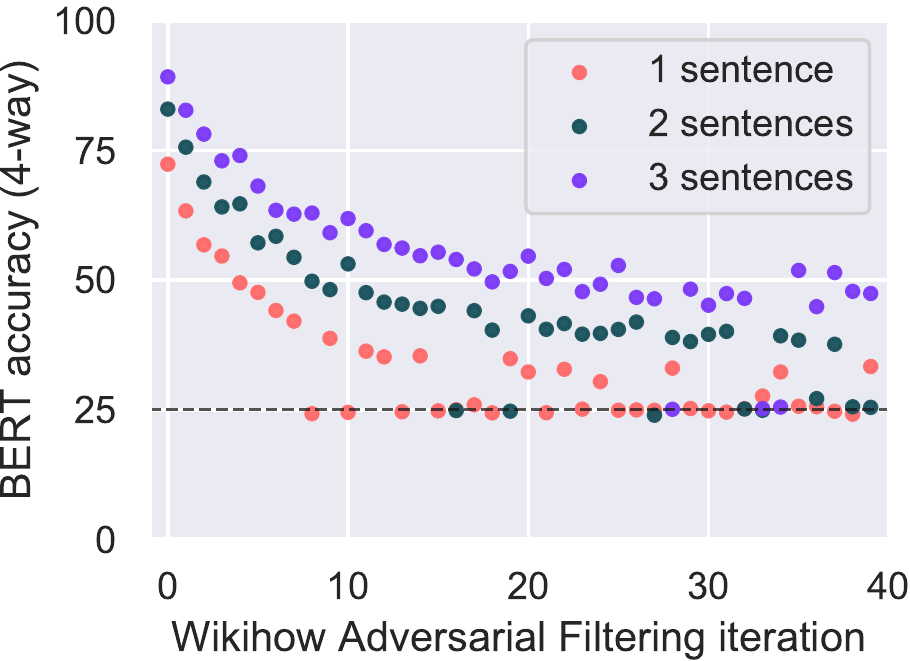}
    
  \caption{Adversarial Filtering (AF) results with BERT-Large as the discriminator. \textbf{Left}: AF applied to ActivityNet generations produced by \citet{zellers2018swagaf}'s language model versus OpenAI GPT. While GPT converges at random, the LM used for \olddatasetname~converges at 75\%. \textbf{Right}: AF applied to WikiHow generations from GPT, while varying the ending length from one to three sentences. They converge to random, $\sim$40\%, and $\sim$50\%, respectively.}
  \label{fig:afs}
\end{figure*}

Figure~\ref{fig:ablations} compares BERT's performance when trained and evaluated on variants of \olddatasetname. 

\noindent \textbf{Context:} BERT's performance only slips 11.9 points (86.7\%$\rightarrow$74.8\%) when context is omitted (\texttt{Ending Only}), suggesting a bias exists in the endings themselves.\footnote{These biases are similar to those in NLI datasets, as found by \citet{gururangan2018annotation, poliak_hypothesis_2018}.}
If a followup event seems unreasonable \emph{absent of context}, then there must be something markedly different between the space of human-written and machine-generated endings. %

\noindent \textbf{Structure:} To distinguish word usage from structural patterns, we consider a new scenario, \texttt{Shuffled}. Here the shared context is provided, but the words in each ending choice are randomly permuted. Surprisingly, this reduces BERT performance by less than 10\%. Even though BERT was never exposed to randomly shuffled text during pretraining, it easily adapts to this setting, which suggests that BERT is largely performing lexical reasoning over each (context, answer) pair. 

Finally, when the context is removed and the words in each ending are shuffled, performance drops to 60.4\%. While low, this is still higher than ELMo's performance (${<}60$\% from \citealp{zellers2018swagaf}). As neither context nor structure is needed to discriminate between human and machine-written endings in a majority of cases, it is likely that systems primarily learn to detect distributional stylistic patterns during finetuning.

\subsection{Where do the stylistic biases come from?}
\olddatasetname~was constructed via Adversarial Filtering (AF). Endings were generated via a language model, and then selected to fool a discriminator. To understand why it was solved requires understanding the interplay of AF with respect to \olddatasetname's generators and discriminators.

\citet{zellers2018swagaf} used a two-layer LSTM for generation, with shallow stylistic adversarial filters.\footnote{The discriminator was an ensemble that featured a bag of words model, a shallow CNN, a multilayer perceptron operating on language model perplexities.} This setup was robust against ELMo models, but has the shallow LM in particular produced distributional artifacts that BERT picks up on?

To investigate this, we perform AF using BERT-Large as the discriminator\footnote{On each iteration, BERT-Large is re-initialized from its pretrained checkpoint, finetuned, and then evaluated in a four-way setting on the dummy test set of held-out data. See Supp \ref{sec:app_AF} for a details of our BERT-Large AF setup.} in two settings, comparing generations from \citet{zellers2018swagaf} with those from a finetuned GPT \cite{radford2018improving}.

Strikingly, the results, Figure~\ref{fig:afs} (left), show that the generations used in \olddatasetname~are so different from the human-written endings that \emph{AF never drops the accuracy to chance}; instead, it converges to roughly 75\%. On the other hand, GPT's generations are good enough that BERT accuracy drops below 30\% over many random subsplits of the data, revealing the importance of the generator.

\section{\Large \newdatasetname}
\label{sec:hellaswag}
The success of BERT implies that high-quality generators and discriminators are crucial to AF's success. However, it does \emph{not} imply that the underlying task of commonsense NLI -- as opposed to a single dataset -- is solved. To evaluate this claim requires us to try making a new evolution of the SWAG dataset, one in which artifacts are removed. In this section, we do just that by introducing \newdatasetname. 
\subsection{ActivityNet Captions}
We start by including video captions from the ActivityNet Captions dataset \cite{krishna_dense-captioning_2017}. The original SWAG dataset contains these, along with captions from LSMDC \cite{rohrbach_movie_2017}, but for \newdatasetname~we solely used ActivityNet. In addition to temporal descriptions, ActivityNet also provides activity labels for each caption (e.g. \texttt{jumping rope}). We will use these activity labels as additional structure to test generalization ability.

\subsection{WikiHow: A New Testbed}
We next consider a new and challenging testbed for commonsense reasoning: completing how-to articles from WikiHow, an online how-to manual. We scrape 80k context and follow-up paragraphs from WikiHow, covering such diverse topics as ``how to make an origami owl'' to ``how to survive a bank robbery.'' Each context has at most three sentences, as do the follow-ups.

AF's effectiveness in this new setting is shown in Figure~\ref{fig:afs} (right). We consider three settings, corresponding to endings that are either one, two, or three sentences long. In all cases, BERT performance begins high (70-90\%), but there are enough generations for Adversarial Filtering to lower the final accuracy considerably. While the one-sentence case converges to slightly higher than random -- 35\% when it converges -- the two and three sentence cases are higher, at 40\% and 50\% respectively. Given more context, it becomes easier to classify an ending as machine- or human-written. We compromise and use two-sentence generations. Particularly in the two-sentence case, we find ourselves in a Goldilocks zone wherein generations are challenging for deep models, yet as we shall soon see, easy for humans.

\subsection{Obtaining high human agreement}
\begin{figure}[t!]
  \centering\small
    \includegraphics[width=\columnwidth]{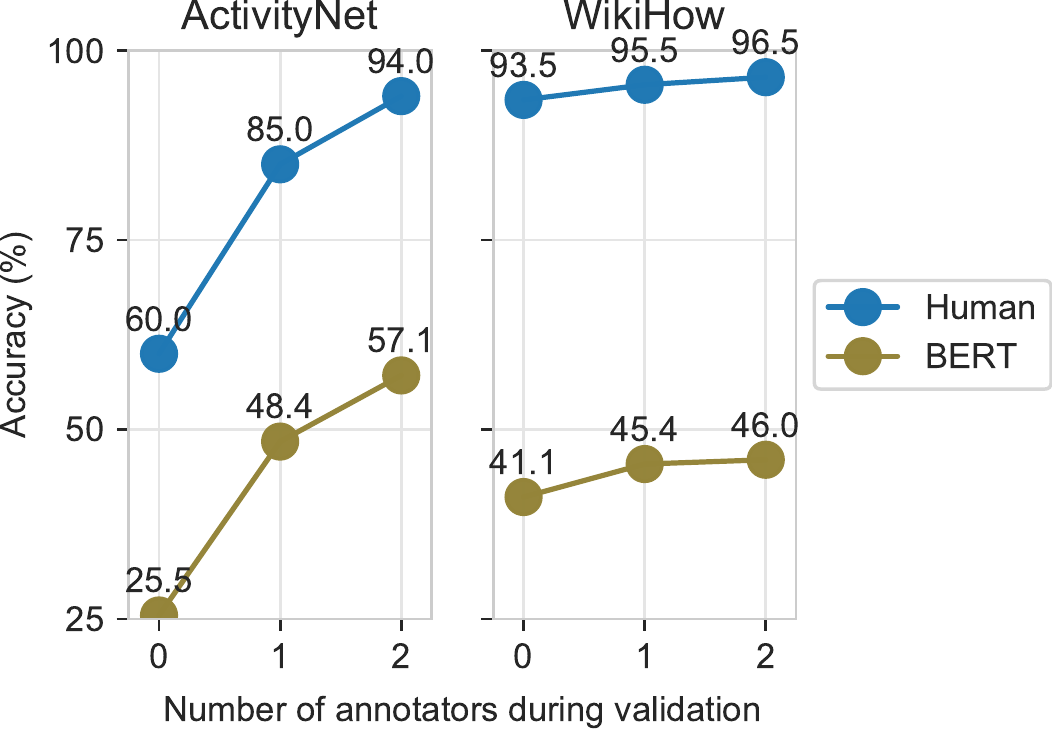}
\caption{For \newdatasetname, we ensure high human agreement through several rounds of annotation. By collecting how likely each ending is we can filter false negative endings -- machine generations that sound realistic -- and replace them with true negatives. On both subdatasets, BERT performance increases during validation, but the gap to human performance remains wide.}
  \label{fig:agreement}
\end{figure}

\begin{figure}[t!]
  \centering\small
    \includegraphics[width=\columnwidth]{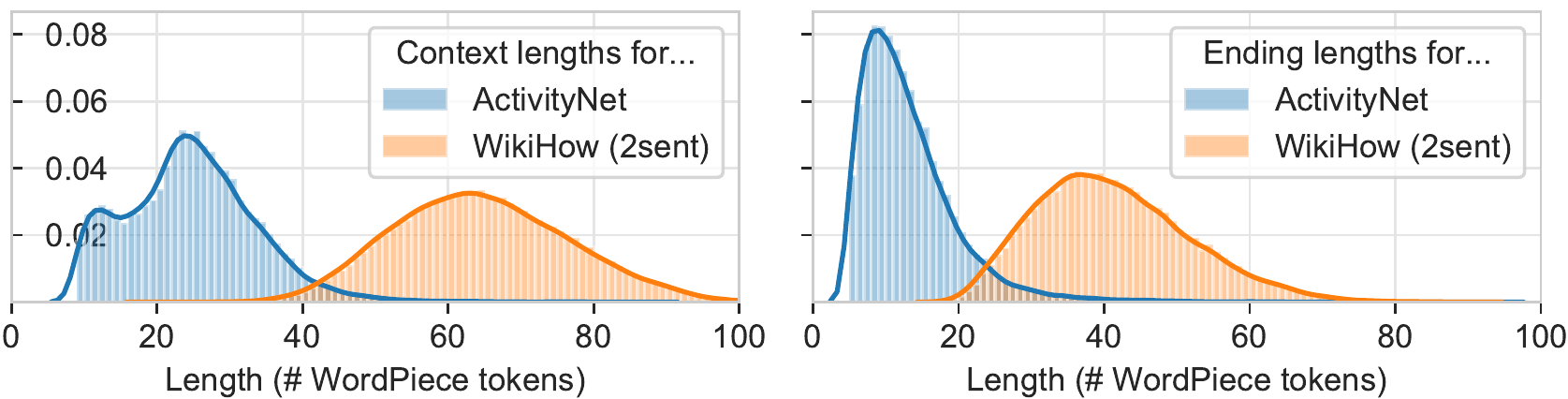}
\caption{Lengths of ActivityNet and WikiHow; the latter with two-sentence generations. WikiHow is much longer, which corresponds to being easier for humans, while taking longer for AF to converge.}
  \label{fig:lengths}
\end{figure}

How well can humans distinguish human-written endings from machine generations refined with Adversarial Filtering? In Figure~\ref{fig:agreement}, we compare human performance with that of BERT on a random 80\%/20\% split. We see a contrast between the ActivityNet and WikiHow performance. While ActivityNet starts off harder for BERT (25.5\%), it also proves difficult for humans (60\%). In contrast, WikiHow starts easier for BERT (41.1\%) and humans find the domain almost trivial (93.5\%).
We hypothesis this discrepancy is due to the lengths of both datasets (Figure~\ref{fig:lengths}). WikiHow's 2-sentence generations average 41 tokens, versus 13 for ActivityNet. This gives WikiHow generations three times as many opportunities to make a detectable mistake. 

To ensure high agreement on ActivityNet, we perform several rounds of human filtering, increasing human performance to 94\%. During human validation, crowd workers are given a context and six ending choices, of which one is the true ending, and the other five are from AF. On each iteration, we replace machine-written endings that the worker rated as realistic with new samples. In the end, we keep the 25k best ActivityNet contexts (i.e. those with highest agreement among workers
\footnote{See the appendix for details about how we estimate this.})
and the 45k best WikiHow contexts. 

\subsection{Zero-shot categories for evaluation}
\begin{figure}[t!]
  \centering\small
    \includegraphics[width=\columnwidth]{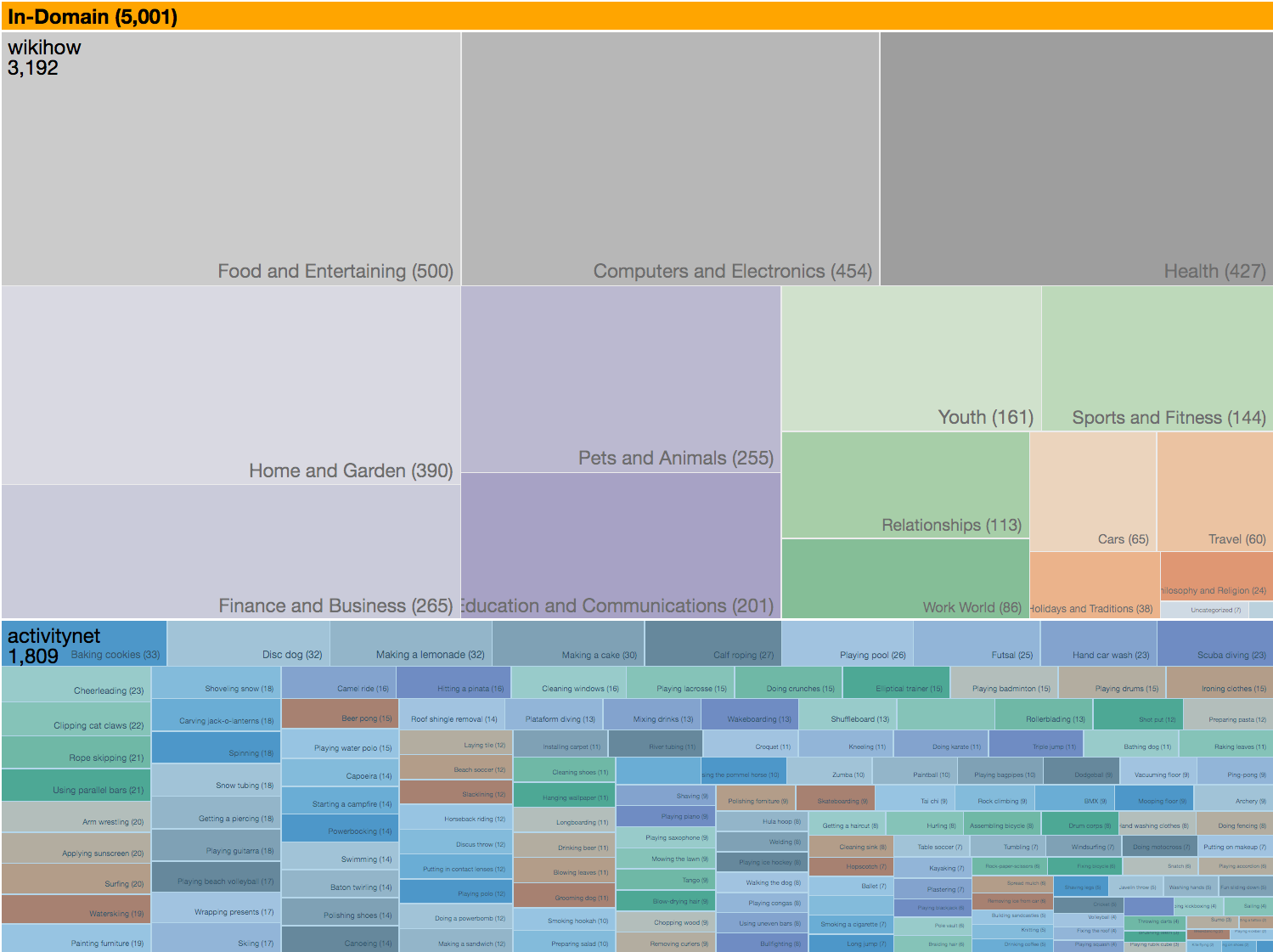}
\caption{Examples on the in-domain validation set of \newdatasetname, grouped by category label. Our evaluation setup equally weights performance on categories seen during training as well as out-of-domain. 
}
  \label{fig:split}
\end{figure}

To evaluate a model's ability to generalize to new situations, we use category labels from WikiHow and ActivityNet to make `zero-shot' evaluation sets. For each set (validation or test), we craft two subsets: one containing 5k `in-domain' examples that come from categories as seen during training (Figure~\ref{fig:split}), and another with 5k `zero-shot' examples from randomly chosen held-out categories. In total, there are 70k dataset examples. 

\definecolor{lightgray}{rgb}{0.95, 0.95, 0.95}
\newcolumntype{g}{>{\columncolor{lightgray}}c}

\newcommand{\best}[1]{\textbf{#1}}
\newcommand{\scnd}[1]{#1}

\begin{table*}[t!]
\centering\small
\begin{tabular}{@{}l@{\hspace{2em}}gg@{\hspace{1em}}||@{\hspace{1em}}cc@{\hspace{1.75em}}cc@{\hspace{1em}}||@{\hspace{1em}}cc@{\hspace{1.75em}}cc@{}}
&\multicolumn{2}{@{}g@{\hspace{2.25em}}}{Overall}
&\multicolumn{2}{@{}c@{\hspace{1.25em}}}{In-Domain}
&\multicolumn{2}{@{}c@{\hspace{1.25em}}}{Zero-Shot}
&\multicolumn{2}{@{}c@{\hspace{1.25em}}}{ActivityNet}
&\multicolumn{2}{@{\hspace{0.25em}}c}{WikiHow}\\
Model &Val &Test&Val&Test&Val&Test&Val&Test&Val&Test\\
 \multicolumn{1}{r}{\small Split Size$\rightarrow$ \hspace{0.5mm}} &\small 10K&\small 10K&\small 5K&\small 5K&\small 5K&\small 5K&\small 3.2K&\small 3.5K&\small 6.8K&\small 6.5K\\
\toprule
Chance & \multicolumn{10}{c}{25.0}   \\
\midrule
fastText&30.9&31.6&33.8&32.9&28.0&30.2&27.7&28.4&32.4&33.3\\
LSTM+GloVe&31.9&31.7&34.3&32.9&29.5&30.4&34.3&33.8&30.7&30.5\\
LSTM+ELMo&31.7&31.4&33.2&32.8&30.4&30.0&33.8&33.3&30.8&30.4 \\
LSTM+BERT-Base&35.9&36.2&38.7&38.2&33.2&34.1&40.5&40.5&33.7&33.8\\
ESIM+ELMo&33.6&33.3&35.7&34.2&31.5&32.3&37.7&36.6&31.6&31.5\\
OpenAI GPT&\scnd{41.9}&\scnd{41.7}&\scnd{45.3}&\scnd{44.0}&\scnd{38.6}&\scnd{39.3}&46.4&43.8&\scnd{39.8}&\scnd{40.5}\\
BERT-Base&39.5&40.5&42.9&42.8&36.1&38.3&\scnd{48.9}&\scnd{45.7}&34.9&37.7\\
BERT-Large&\best{46.7}&\best{47.3}&\best{50.2}&\best{49.7}&\best{43.3}&\best{45.0}&\best{54.7}&\best{51.7}&\best{42.9}&\best{45.0}\\
\midrule
Human&95.7&95.6&95.6&95.6&95.8&95.7&94.0&94.0&96.5&96.5\\
\bottomrule
\end{tabular}
\vspace*{-1mm}\caption{Performance of models, evaluated with accuracy (\%).We report results on the full validation and test sets (Overall), as well as results on informative subsets of the data: evaluated on in-domain, versus zero-shot situations, along with performance on the underlying data sources (ActivityNet versus WikiHow). All models substantially underperform humans: the gap is over 45\% on in-domain categories, and 50\% on zero-shot categories.
}\vspace*{-1mm}

\label{tab:results}
\end{table*}

\section{Results}
We evaluate the difficulty of \newdatasetname~using a variety of strong baselines, with and without massive pretraining. The models share the same format: given a context and an ending, return a \emph{logit} for that ending. Accordingly, we train our models using a four-way cross-entropy loss, where the objective is to predict the correct ending. In addition to BERT-Large, our comparisons include:
\begin{enumerate}[wide, labelwidth=!,labelindent=0pt,noitemsep,topsep=0pt,label=\textbf{\alph*}.]
\item {\bf OpenAI GPT} \cite{radford2018improving}: A finetuned 12-layer transformer that was pre-trained on the BookCorpus \cite{moviebook}.
\item {\bf Bert-Base}: A smaller version of the BERT model whose architecture size matches GPT.
\item {\bf ESIM+ELMo} \cite{chen2017enhanced, peters2018deep}: This is the best-performing ELMo model for NLI, modified slightly so the final output layer is now a four-way softmax over endings.
\item {\bf LSTM sentence encoder}: This is a randomly initialized two-layer bi-LSTM; the second layer's hidden states are max-pooled and fed into an MLP to predict the logit. We consider three variations: GloVe embeddings, ELMo embeddings, or (frozen) BERT-Base embeddings.\footnote{For ELMo and BERT-Base, the model learns scalar weights to combine each internal layer of the encoder.}
\item {\bf FastText}: \cite{joulin2017bag} An off-the-shelf library for bag-of-words text classification.\footnote{This model is trained with binary cross entropy loss.}
\end{enumerate}

We compare all models to human performance by asking five independent crowd workers to solve the same four-way multiple choice problems; their predictions are combined via majority vote.

Our results, shown in Table~\ref{tab:results}, hint at the difficulty of the dataset: human performance is over 95\%, while overall model performance is below 50\% for every model. Surprisingly, despite BERT-Large having been used as the adversarial filter, it still performs the strongest at 47.3\% overall. By making the dataset adversarial for BERT, it seems to also have become adversarial \textbf{for every other model.} For instance, while ESIM+ELMo obtained 59\% accuracy on \olddatasetname, it obtains only 33.3\% accuracy on \newdatasetname.

In addition to pretraining being critical, so too is end-to-end finetuning. Freezing BERT-Base and adding an LSTM on top lowers its overall performance 4.3\%. This may help explain why models such as ESIM+ELMo struggled on \olddatasetname, as ELMo isn't updated during finetuning.

While BERT is the best model, it still struggles on \newdatasetname, and especially so on zero-shot categories. Performance drops roughly 5\% on the test fold, which suggests that the finetuning is not enough for BERT to learn to generalize to novel activities or how-to categories.

Last, we see that WikiHow is a much harder domain that ActivityNet for machines: 45\% Bert-Large performance, versus 96.5\% for humans. Curiously, it is on this source dataset that we see the smallest gap between OpenAI GPT and BERT. In fact, OpenAI GPT outperforms BERT on WikiHow, but the reverse is true for ActivityNet. One possibility is that the left-to-right structure of GPT is the right inductive bias for WikiHow - perhaps reasoning bidirectionally over long contexts is too much for a 12-layer transformer to learn.

\subsection{\olddatasetname~to \newdatasetname~transfer}
\begin{figure}[t!]
  \centering\small
    \includegraphics[width=\columnwidth]{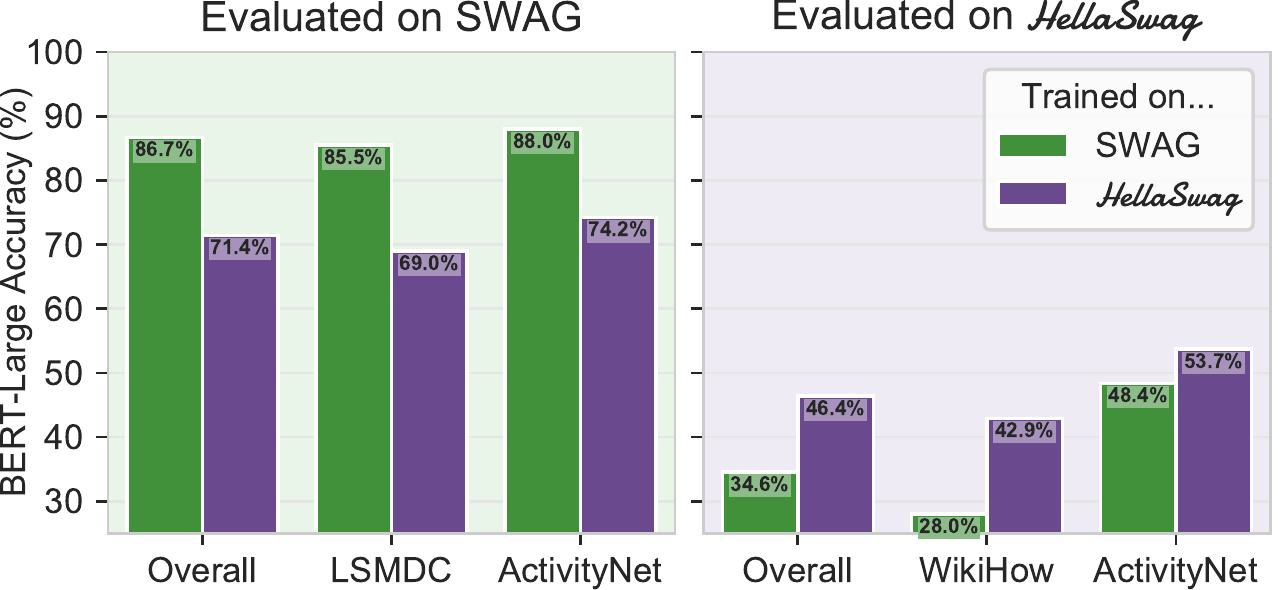}
\caption{Transfer experiments from \olddatasetname~to \newdatasetname~and vice versa, evaluated on the validation sets. Overall, a BERT-Large that is trained on \olddatasetname~hardly generalizes to \newdatasetname: it scores 34.6\%. }
  \label{fig:swagstransfer}
\end{figure}

Given the shared goals and partial domains of \olddatasetname~and \newdatasetname, it is natural to ask to what extent models can transfer between the two datasets.
In Figure~\ref{fig:swagstransfer} we show the results from transfer experiments: models are trained on one dataset and evaluated on the other.\footnote{Note that the ActivityNet splits are different for each dataset. To avoid skewing the results, we report only on the validation video captions that are not in the training sets of either dataset. The overall accuracy is then a weighted average, where ActivityNet examples are weighted proportionately more. This gives a slight advantage to training on \olddatasetname, as it sees all the ActivityNet categories when training.} 

The best models are trained on the same dataset that they are evaluated on: training on \olddatasetname~and evaluating on \newdatasetname~lowers performance by 12\%; vice versa lowers performance by 15\%. The missing domain for \newdatasetname~models is movie descriptions (LSMDC), still, \newdatasetname~models obtain 69\% accuracy. On the other hand, \olddatasetname~models do not generalize at all to their missing domain, WikiHow (28\%), suggesting that learning general commonsense reasoning was hardly necessary to solve \olddatasetname.

\subsection{Qualitative examples}
We show several qualitative examples in Figure~\ref{fig:tinyexamples}, along with BERT-Large's predictions. BERT does well on some ActivityNet contexts, such as in the first row, where it correctly predicts the ending for a \texttt{shaving} caption. Whereas \emph{shaving} is in-domain, the second example about \texttt{sharpening knives} is zero-shot. In this context, BERT's answer suggests that one would use a knife to sharpen a stone, rather than vice versa. The last example comes from WikiHow, which appears to be incredibly challenging for BERT. BERT picks answer \textbf{d}, which has more words that match the context of \emph{technology} (planes, traffic, laptop), but is incoherent.\footnote{Among other issues, why would someone suddenly be aware that they are `flying at high speed on a plane...?'}

\begin{figure}[t]
\centering\footnotesize
    {\FrameSep1pt
    \begin{framed}\scriptsize\begin{tabular}{@{}l @{}}
\aquestion{\textbf{Category}: Shaving (ActivityNet; In-domain)}{A bearded man is seen speaking to the camera and making several faces. the man}{a) then switches off and shows himself via the washer and dryer rolling down a towel and scrubbing the floor. (0.0\%)}{b) then rubs and wipes down an individual's face and leads into another man playing another person's flute. (0.0\%)}{c) is then seen eating food on a ladder while still speaking. (0.0\%)}{\correctans{d) then holds up a razor and begins shaving his face. (100.0\%)}} \\ \midrule
\aquestion{\textbf{Category}: Sharpening knives (ActivityNet; Zero-Shot)}{Two men are in a room and the man with a blue shirt takes out a bench stone and with a little lubricant on the stone takes an knife and explains how to sharpen it. then he}{\incans{a) uses a sharpener to smooth out the stone using the knife. (100.0\%)}}{b) shows how to cut the bottom with the knife and place a tube on the inner and corner. (0.0\%)}{c) bends down and grabs the knife and remove the appliance. (0.0\%)}{\missed{d) stops sharpening the knife and takes out some pieces of paper to show how sharp the knife is as he cuts slivers of paper with the knife. (0.0\%)}}  \\ \midrule
\aquestion{\textbf{Category}: Youth (WikiHow; In-Domain)}{\textsc{How to make up a good excuse for your homework not being finished} \\ \\ \textbf{Blame technology.}  One of the easiest and most believable excuses is simply blaming technology. You can say your computer crashed, your printer broke, your internet was down, or any number of problems.}{a) Your excuses will hardly seem believable. {[}substeps{]} This doesn't mean you are lying, just only that you don't have all the details of how your computer ran at the time of the accident. (0.0\%)}{b) The simplest one to have in a classroom is to blame you entire classroom, not just lab. If you can think of yourself as the victim, why not blame it on technology. (9.4\%)}{\missed{c) Most people, your teacher included, have experienced setbacks due to technological problems. {[}substeps{]} This is a great excuse if you had a paper you needed to type and print. (29.1\%)}}{\incans{d) It may also be more believable if you are fully aware that you may be flying at high speed on a plane and need someone to give you traffic report. Your problem might be your laptop failing to charge after a long flight. (61.5\%)}}
\end{tabular}
\end{framed} }
\vspace*{-3mm}\caption{Example questions answered by BERT-Large. Correct model predictions are \correctans{blue}, incorrect predictions are \incans{red}. The right answers are \textbf{bolded}.}
\label{fig:tinyexamples}
\end{figure}

\section{Discussion}
Our results suggest that \newdatasetname~is a challenging testbed for state-of-the-art NLI models, even those built on extensive pretraining. The question still remains, though, of \emph{where will the field go next?}

\begin{figure}[t!]
  \centering\small
    \includegraphics[width=\columnwidth]{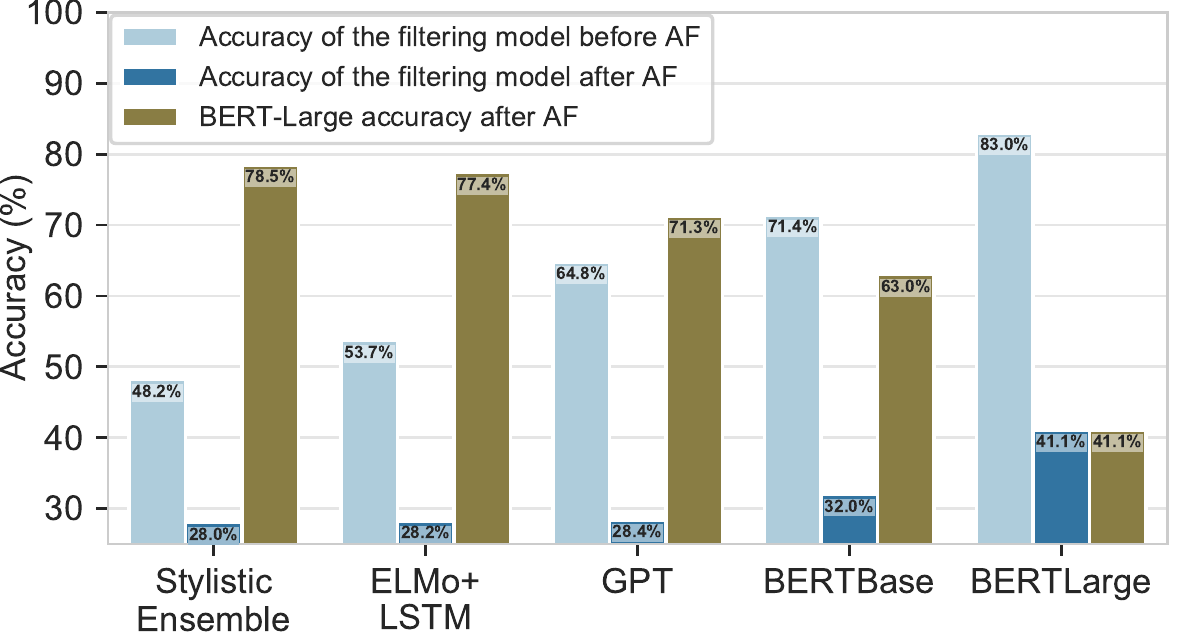}
\caption{Performance on the WikiHow subset of alternative variations of \newdatasetname, where different Adversarial Filters are used (but without human validation). We consider the shallow stylistic adversaries used by \citet{zellers2018swagaf} (Stylistic Ensemble), as well as an LSTM with ELMo embeddings, GPT, BERT-Base, and BERT-Large. For each adversarial filtering model, we record the accuracy of that model before and after AF is used. We also evaluate each alternative dataset using BERT-Large. The results suggest that using a a stronger model at test time (over the model used for AF) improves performance, but is not enough to solve the task.
}
  \label{fig:varydisc}
\end{figure}
\subsection{How easy might \newdatasetname~be for future discriminators?}
In this paper, we showed the existence of a Goldilocks zone of text complexity -- in which generations are nonsensical, but existing state-of-the-art NLP models cannot tell the difference. How hard will the dataset be for future, even more powerful, models? 

Answering this question is challenging because \emph{these models don't exist (or are unavailable) at the time of writing}. However, one remedy is to perform an ablation study on the Adversarial Filtering model used, comparing weaker filters with stronger discriminators. We present our results in Figure~\ref{fig:varydisc}, and find that while weak discriminators (like the stylistic ensemble used to make SWAG) only marginally reduce the accuracy of BERT-Large, increasing the gap between the filter and the final discriminator is not enough to solve the task. For instance, using a discriminator with 3x the parameters as the adversarial filter (BERT-Large vs. BERT-Base) results in 63\% machine accuracy.

\subsection{How well does pretraining scale?}
\begin{figure}[t!]
  \centering\small
    \includegraphics[width=\columnwidth]{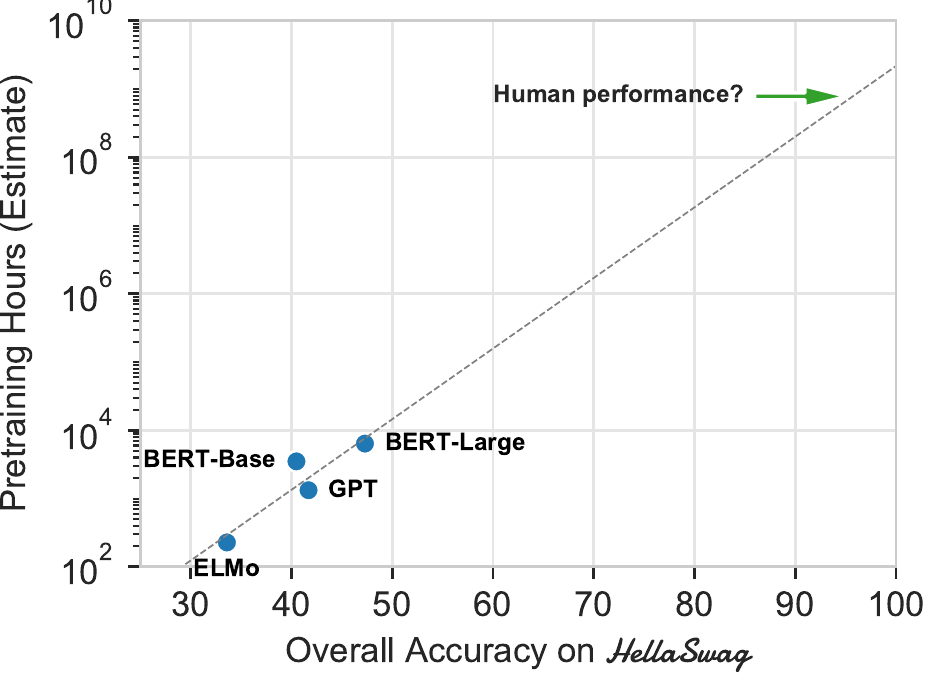}
\caption{Estimated pretraining hours required to reach a desired accuracy on \newdatasetname. We estimate perfomance with respect to a RTX 2080 Ti - a modern, fast GPU, and fit a log-linear regression line. An extrapolation suggests that to reach human-level performance on \newdatasetname, without algorithmic or computational improvements, would require $10^9$ GPU-hours of pretraining (over 100k GPU years).}
  \label{fig:GPU}
\end{figure}
Overall, the current paradigm of pretraining large models on lots of data has made immense progress on NLP benchmarks. Though we expect this trend to continue, it also behooves us to consider its limits. If more compute is indeed the answer for human-level commonsense inference, what would the compute requirements of this hypothetical massive model look like?

We investigate this in Figure~\ref{fig:GPU} by comparing the accuracies of known models on \newdatasetname~with their computational needs. This estimation is a rough estimate: we convert reported TPU runtimes to our benchmark RTX 2080 Ti GPU using the Roofline model \cite{williams2009roofline}, which focuses primarily on the bottleneck of loading tensors into GPU memory. Extrapolating from an exponential fit suggests that reaching human-level performance on our dataset would require $10^9$ GPU hours, or 100k years -- unless algorithmic improvements are made.

What might these algorithmic improvements look like? These could include architectural advances, better pretraining objectives, and beyond. However, these improvements share the bottleneck of
the data source. To answer some \newdatasetname~questions correctly without reasoning deeply -- like knowing that it is a bad idea to stop at a red light for `at most two seconds' -- might require an exponential number of samples, due to problems of reporting bias \cite{gordon2013reporting}. Alternatively, future models might answer correctly only by picking up on spurious patterns, in which case a new development of the benchmark -- using these models as adversaries -- would place us in the same position as we are right now.

Put another way, for humans to answer \newdatasetname~questions requires \emph{abstracting away} from language and modeling \emph{world states} instead. We postulate that this is what separates solving the \emph{task} of commonsense NLI, as opposed to a particular dataset. Indeed, we find that existing deep methods often get fooled by lexical false friends. For example, in the WikiHow example from Figure~\ref{fig:tinyexamples}, BERT chooses an ending that matches the \emph{technology} words in the context, rather than matching the deeper topic: using technology as an excuse for not doing homework. 

\subsection{Towards a future of evolving benchmarks}
What happens when \newdatasetname~gets solved? We believe the answer is simple: crowdsource another dataset, with the same exact format, and see where models fail. Indeed, in our work we found this to be straightforward from an \emph{algorithmic} perspective: by throwing in the \emph{best known generator} (GPT) and the \emph{best known discriminator} (BERT-Large), we made a dataset that is adversarial - not just to BERT, but to all models we have access to.

While this was easy algorithmically, care must be taken from a data curation standpoint. Indeed, we find success exists within a Goldilocks zone: the data source must be complex enough that state-of-the-art generators often make mistakes, while simple enough such that discriminators often fail to catch them. This ties the future of \olddatasetname-style benchmarks to progress on language generation: until generation is solved, commonsense NLI will remain unsolved. Even recent promising results on scaling up language models \cite{radford2019gpttwo} find problems in terms of consistency, with the best curated examples requiring 25 random seeds.

\section{Conclusion}
In this paper, we presented \newdatasetname, a new dataset for physically situated commonsense reasoning. By constructing the dataset through adversarial filtering, combined with state-of-the-art models for language generation and discrimination, we produced a dataset that is adversarial to the most robust models available -- even when models are evaluated on items from the training distribution. In turn, we provided insight into the inner workings of pretrained models, and suggest a path for NLP progress going forward: towards benchmarks that adversarially co-evolve with evolving state-of-the-art models.

\section*{Acknowledgments}
We thank the reviewers, as well as Jesse Thomason, for their helpful feedback. We thank the Mechanical Turk workers for their great work during dataset collection. Thanks also to Zak Stone and the Google Cloud TPU team for help with the computing infrastructure. This work was supported by the National Science Foundation through a Graduate Research Fellowship (DGE-1256082) and NSF grants (IIS-1524371, 1637479, 165205, 1703166), the DARPA CwC program through ARO (W911NF-15-1-0543), the IARPA DIVA program through D17PC00343, the Sloan Research Foundation through a Sloan Fellowship, the Allen Institute for Artificial Intelligence, the NVIDIA Artificial Intelligence Lab, and gifts by Google and Facebook. The views and conclusions contained herein are those of the authors and should not be interpreted as representing endorsements of IARPA, DOI/IBC, or the U.S. Government. 

\bibliography{hellaswag}
\bibliographystyle{acl_natbib}

\appendix
\clearpage
\section*{Supplemental Material}
\section{Adversarial Filtering Setup}
\label{sec:app_AF}
In this subsection, we provide some more details regarding the Adversarial Filtering experiments.  

Our version of Adversarial Filtering is mostly the same as \citet{zellers2018swagaf}. Details:
\begin{enumerate}[wide, labelwidth=!,labelindent=0pt,noitemsep,topsep=0pt,leftmargin =*,label=\textbf{\alph*}.]
\item On each iteration, we split the dataset up into 80\% training and 20\% testing. We don't do anything special for this split (like looking at the video/article IDs).
\item For ActivityNet, we use $k=9$ assigned indices for every example. (This corresponds to the number of red columns in Figure 2). For WikiHow, we used $k=5$, since we found that there were fewer good endings produced by the generators after scaling up the sequence length.
\item Similarly to \citet{zellers2018swagaf}, we train the AF models in a multi-way fashion. Since we use BERT-Large as the discriminator, this matches \citet{devlin2018bert}'s model for \olddatasetname: on each training example, the model is given exactly one positive ending and several negative endings, and the model computes probability distribution over the endings through a softmax. However, we also wanted to always report 4-way probability for simplicity. To do this, we train in a 4-way setting (the training set is constructed by subsampling $3$ wrong answers from the set of $k$ that are currently assigned to each example). The accuracy values that are reported are done so using the first 3 assigned negatives in dataset $\mathcal{D}_{test}$.
\item Sometimes, BERT never converges (accuracy around 25\%), so when this happens, we don't do the reassignment.

\end{enumerate}

\section{GPT Setup}
We generate our dataset examples from OpenAI GPT. We finetune the model for two epochs on WikiHow, and 5 epochs on ActivityNet, using the default learning rate of \cite{radford2018improving}. Importantly, we generate randomly according to the language model distribution, rather than performing beam search -- this would bias the generations towards common words. For the WikiHow endings, we used Nucleus Sampling with $p=0.98$, which means that the probability weights for the tail (those tokens with cumulative probability mass $<0.02$) are zeroed out \citep{holtzman2019curious}.

\section{BERT setup}
We extensively study BERT in this paper, and make no changes to the underlying architecture or pretraining. For all of the experiments where we provide context, we set up the input to the BERT model like this:

{\small \texttt{[CLS] A woman is outside with a bucket and a dog. The dog is running around trying to avoid a bath. [SEP] She gets the dog wet, then it runs away again [SEP]}}

In the case where only the ending is provided, we adopt the BERT-style `single-span' setting:
{\small \texttt{[CLS] She gets the dog wet, then it runs away again [SEP]}}

\section{A discussion on BERT Hyperparameters and Instability}
It is worth noting that many of our experiments some instability. On the \olddatasetname~experiments, we use the same hyperparameters as \cite{devlin2018bert} - these generally work very well.\footnote{The only exception is for the plots where we vary the number of training examples. In this case, we don't want to disadvantage the trials without much training data (since this would allow for fewer parameter updates). To remedy this, we continue training for 10 epochs and report the best validation performance over the entire training history.} However, we find that they become a bit unstable when crossing over to make \newdatasetname. Here, we discuss some strategies and insight that we picked up on.

\begin{enumerate}[wide, labelwidth=!,labelindent=0pt,noitemsep,topsep=0pt,leftmargin =*,label=\textbf{\alph*}.]
\item We use a batch size of 64 examples rather than 16, and warm the model up for 20\% of the dataset (rather than 10\%). This helps the model adapt to SWAG more gradually, without diverging early on.
\item For the Adversarial Filtering experiments (for both WikiHow and ActivityNet), we randomize some of the hyperaparmeters on each iteration. We sample a learning rate between \texttt{1e-5} and \texttt{4e-5}, using a log-uniform distribution. These outer ranges were recommended from the original BERT paper. Additionally, with probability $0.5$ we use the cased model (where the input isn't originally lowercased before tokenization), rather than the uncased model.
\item During adversarial filtering, we used 3 epochs. However, we found that adding more epochs helped the model during fine-tuning on the final dataset \newdatasetname. Our best configuration uses 10 epochs.
\item While fine-tuning on \newdatasetname~we used a learning rate of \texttt{2e-5}.
\end{enumerate}

\section{Human validation}
\label{sec:app_human_val}
We performed human validation using the same setup as \cite{zellers2018swagaf}. Humans get six answers to choose from, of which exactly one is the true ending and the other five are from AF. We found that multiple rounds of human validation were especially helpful on ActivityNet. However, it helps to do the human validation in an intelligent way: if the first worker is confused, the answer should be replaced before it goes to the next worker. This is a hard problem, so we adopt the following approach:

\begin{enumerate}[wide, labelwidth=!,labelindent=0pt,noitemsep,topsep=0pt,leftmargin =*,label=\textbf{\alph*}.]
    \item We use best practices on mechanical turk, paying workers fairly (up to 37 cents per HIT on WikiHow). We also used a qualification HIT that was autograded to help filter for workers who are good at the task. Workers who tended to prefer the generated endings over the real ones were dequalified from participating.
    \item For each worker, we use the summary of their performance so far to estimate $P(\textrm{answer }i \textrm{ is right} | \textrm{worker rates } i \textrm{ as best})$. We can then use this to estimate how confident we are in each answer choice: we want to be confident that workers will \emph{not} prefer the wrong answers. Also, this allows us to aggregate performance across crowd workers, by multiplying the probabilities for each answer choice.
    \item On each round of filtering, we \emph{keep} the 3 wrong endings that workers least prefer (based on the probability scores, along with the right ending. The other two endings are new ones.
\end{enumerate}

Particularly on ActivityNet, we found that there are some contexts where the ground truth answer isn't liked by workers. To fix this, we end up taking the best 25k examples from ActivityNet and the best 45k from WikiHow. (By best, we mean the ones with the highest probability that workers will predict the true answer, versus the three easiest-to-guess negatives, as judged by the Naive Bayes model). We make Figure 7 (`The road to \newdatasetname') by doing this process (taking the best examples) for each dataset, while varying the number of annotators that are used for getting the scores for each ending. (In the case where there are 0 annotators, we get a random sample). 

\section{Human Evaluation}
\label{sec:app_human_eval}
We do a human evaluation while giving workers the exact same task as is given to the models. Workers are given five endings, and must pick the best one. We obtain human evaluation numbers by combining 5 turkers together, with a majority vote.

We found that the biggest differences in difficulty in humans were due to domain (WikiHow is easier than ActivityNet). To account for this, we did the human evaluation over 200 examples from WikiHow, and 200 examples from ActivityNet, for each number of previous validators as shown in Figure 7 (0, 1, or 2). To report the accuracy of a split that's mixed between WikiHow and ActivityNet, we use the following formula:
\[
\frac{acc_{WikiHow}\cdot N_{WikiHow} + acc_{ActivityNet}\cdot N_{ActivityNet}}{N_{WikiHow} + N_{ActivityNet}}
\]
Here, $acc$ refers to the accuracy on each dataset as judged by humans, and $N$ is the number of examples from that dataset in the split.

\section{More examples}

\begin{table*}[t]
\centering\small
\begin{tabular}{@{}l @{\hspace{0.1cm}}|@{\hspace{0.1cm}} l@{}}
\aquestion{Category: Preparing pasta (activitynet; indomain)}{A kitchen is shown followed by various ingredients and a woman speaking to the camera. She begins showing the ingredients and putting them into a hot boiling pot and stirring around. she}{a) shows off the oven and begins assembling the cookies in the oven by pushing a button on the oven. (2.2\%)}{\correctans{b) continues mixing up more ingredients and then puts them all together in a bowl, serving the dish ad sprinkling olive oil around it. (97.8\%)}}{c) shows raising and lowering the pot until adding more water and corn syrup. (0.0\%)}{d) places an omelette onto the screen and puts it in the oven to bake. (0.0\%)}
 &
\aquestion{\textbf{Category}: Doing crunches (activitynet; indomain)}{We see a fitness center sign. We then see a man talking to the camera and sitting and laying on a exercise ball. the man}{a) demonstrates how to increase efficient exercise work by running up and down balls. (0.0\%)}{\incans{b) moves all his arms and legs and builds up a lot of muscle. (80.9\%)}}{c) then plays the ball and we see a graphics and hedge trimming demonstration. (0.0\%)}{\missed{d) performs sits ups while on the ball and talking. (19.1\%)}} \\ \midrule
\aquestion{Category: Sharpening knives (activitynet; zeroshot)}{A man is seen spinning a blade with his foot on a machine and moving his hands up with down holding a knife. the camera}{a) pans around and shows a woman moving around in a jump rope machine. (0.0\%)}{\correctans{b) captures him from several angles while he sharpens the knife with complete concentration. (81.6\%)}}{c) pans around and points to a man standing inside the machine as the man continues to move on the machine. (18.4\%)}{d) then pans around to a woman and her daughter who also dance at the show. (0.0\%)}
 &
\aquestion{Category: Layup drill in basketball (activitynet; zeroshot)}{A female basketball coach is seen instructing a group of girl basketball players who are standing in line on a basketball court. the first girl}{\missed{a) passes to another coach and then runs to the net and takes a layup. (0.0\%)}}{\incans{b) trying to get the ball to go far past the basket and hit it back towards the basket while her coach continues teaching her. (100.0\%)}}{c) walks across the court with the ball and keeps walking then pulling the girls to the other side of the court and the girls begin playing volleyball rhythmically rolling on the floor as the coach helps them follow how to properly do things. (0.0\%)}{d) line up and stand behind a dummy dummy. (0.0\%)}
 \\  \midrule
\aquestion{Category: Youth (wikihow; indomain)}{{[}header{]} How to make up a good excuse for your homework not being finished {[}title{]} Blame technology. {[}step{]} One of the easiest and most believable excuses is simply blaming technology. You can say your computer crashed, your printer broke, your internet was down, or any number of problems.}{a) Your excuses will hardly seem believable. {[}substeps{]} This doesn't mean you are lying, just only that you don't have all the details of how your computer ran at the time of the accident. (0.0\%)}{b) The simplest one to have in a classroom is to blame you entire classroom, not just lab. If you can think of yourself as the victim, why not blame it on technology. (9.4\%)}{\missed{c) Most people, your teacher included, have experienced setbacks due to technological problems. {[}substeps{]} This is a great excuse if you had a paper you needed to type and print. (29.1\%)}}{\incans{d) It may also be more believable if you are fully aware that you may be flying at high speed on a plane and need someone to give you traffic report. Your problem might be your laptop failing to charge after a long flight. (61.5\%)}}
 & 
\aquestion{Category: Family Life (wikihow; zeroshot)}{{[}header{]} How to raise your children to be helpers {[}title{]} Call them helpers when you ask for things. {[}step{]} Instead of asking for help, ask your child to " be a helper. " all people, children included, are more motivated when their identity is in play.}{\missed{a) You can start doing this with your children as early as two years old. {[}substeps{]} You might say, " jayden, can you be a helper and clean your bedroom before grandma comes over? " or " please be a helper and stay quiet while your sister naps. (0.1\%)}}{\incans{b) When you call your child helpers, describe what they do and what they need to be helped for. {[}substeps{]} You could say, " i need you to help dad during his lunch break at work. (99.9\%)}}{c) If you ask your child for things they have access to, it encourages them to put more effort into making things happen. {[}substeps{]} To make sure they understand exactly what's expected of them, you could try saying, " i'm looking for helpers who can be helpers. (0.0\%)}{d) Call them when you need them for help or for monetary help. {[}substeps{]} For example, if you need help with something you don't know how to do, let your child know you're excited to help with this. (0.0\%)}
\end{tabular}
\vspace*{-3mm}\caption{Example questions answered by BERT-Large. Correct model predictions are in \correctans{blue}, incorrect model predictions are \incans{red}. The right answers are \textbf{bolded}.}
\label{fig:qualitative}
\end{table*}
We additionally have more validation examples, shown in Figure~\ref{fig:qualitative}.

\section{In-Domain and Zero-Shot categories}
\begin{figure*}[t!]
    \includegraphics[width=\textwidth]{figures/indomain.png}\\
    \includegraphics[width=\textwidth]{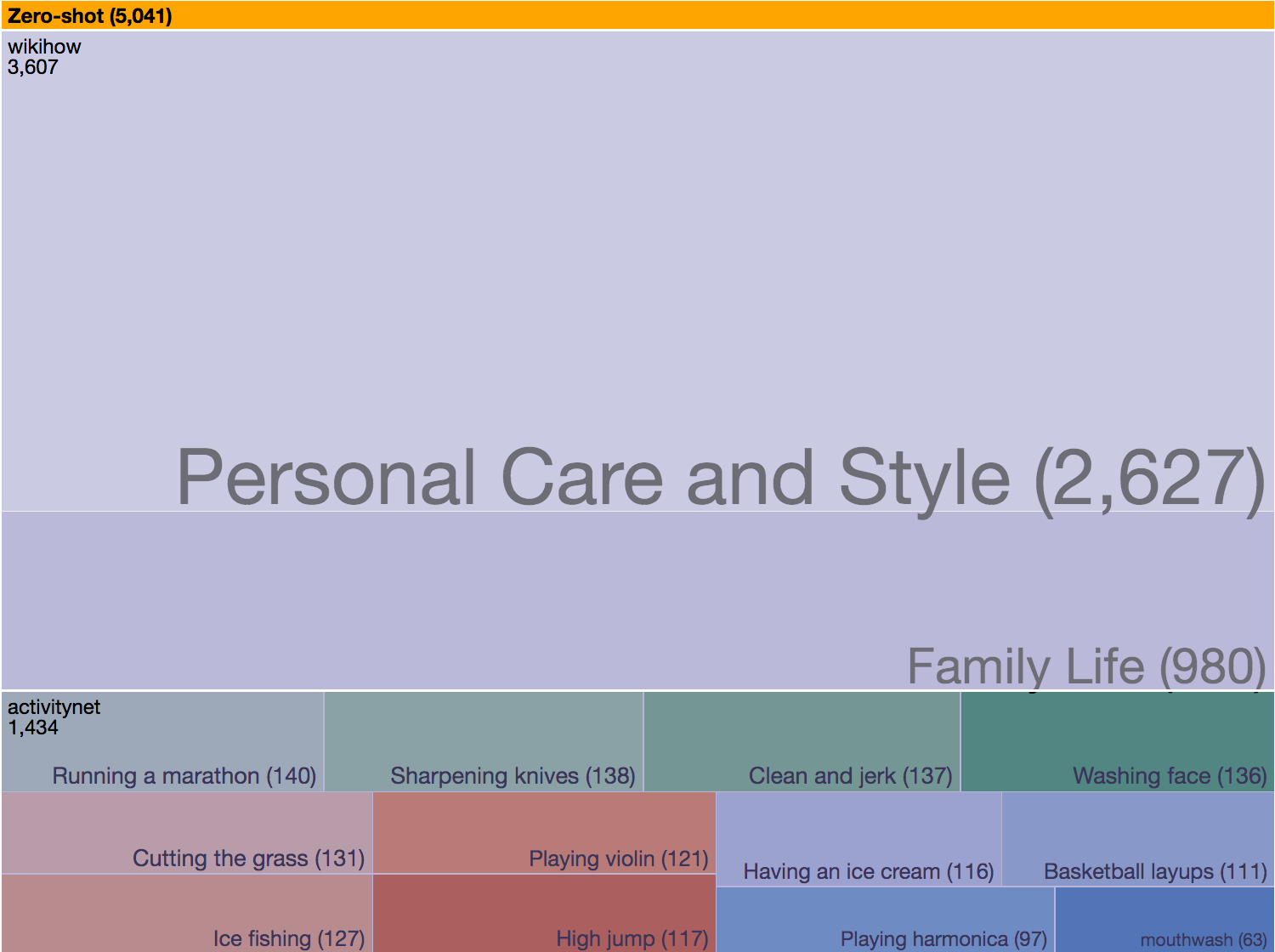}
\caption{Examples on the in-domain validation set of \newdatasetname, grouped by category label. Our evaluation setup equally weights performance on categories seen during training as well as out-of-domain.}
  \label{fig:big_split}
\end{figure*}

See Figure~\ref{fig:big_split} for a closer look at the dataset categories.

\end{document}